\documentclass{article}

\usepackage{PRIMEarxiv}

\usepackage[utf8]{inputenc} 
\usepackage[T1]{fontenc}    
\usepackage{hyperref}       
\usepackage{url}            
\usepackage{booktabs}       
\usepackage{amsfonts}       
\usepackage{nicefrac}       
\usepackage{microtype}      
\usepackage{lipsum}
\usepackage{fancyhdr}       
\usepackage{graphicx}       
\graphicspath{{media/}}     
\usepackage{xcolor}         
\usepackage{makecell}
\newcolumntype{C}[1]{>{\centering\arraybackslash}p{#1}}  
\usepackage{graphicx}
\usepackage{multirow}
\usepackage{makecell}
\usepackage{booktabs}
\usepackage{colortbl}
\usepackage[linesnumbered,ruled,vlined]{algorithm2e}
\usepackage{setspace}
\usepackage{amsmath}
\usepackage{changepage}

\usepackage{textcomp}
\usepackage{stfloats}
\usepackage{url}
\usepackage{caption}
\usepackage{verbatim}
\usepackage{graphicx}
\usepackage{multirow}
\usepackage{makecell}
\usepackage{setspace}
\usepackage{ulem}
\usepackage{colortbl}
\usepackage{xcolor}
\usepackage{booktabs}
\usepackage{pifont}
\definecolor{c2}{HTML}{EBEBEA}

\title{Efficient Text-driven Motion Generation via Latent Consistency Training
}

\author{
  Mengxian Hu \\
  Tongji university \\
  humengxian@tongji.edu.cn \And
  Minghao Zhu \\
  Tongji university \\
  zmhh\_h@126.com \And
  Xun Zhou \\
  Tongji university \\
  tjzhouxun@163.com \And
  Qingqing Yan \\
  Tongji university \\
  qyan\_0131@tongji.edu.cn \And
  Shu Li \\
  Tongji university \\
  lishu@tongji.edu.cn \And
  Chengju Liu \\
  Tongji university \\
  liuchengju@tongji.edu.cn \And
  Qijun Chen \\
  Tongji university \\
  qjchen@tongji.edu.cn
}

\begin{document}
\maketitle

\begin{abstract}
Text-driven human motion generation based on diffusion strategies establishes a reliable foundation for multimodal applications in human-computer interactions.
However, existing advances face significant efficiency challenges due to the substantial computational overhead of iteratively solving for nonlinear reverse diffusion trajectories during the inference phase.
To this end, we propose the motion latent consistency training framework (MLCT), which precomputes reverse diffusion trajectories from raw data in the training phase and enables few-step or single-step inference via self-consistency constraints in the inference phase.
Specifically, a motion autoencoder with quantization constraints is first proposed for constructing concise and bounded solution distributions for motion diffusion processes.
Subsequently, a classifier-free guidance format is constructed via an additional unconditional loss function to accomplish the precomputation of conditional diffusion trajectories in the training phase.
Finally, a clustering guidance module based on the K-nearest-neighbor algorithm is developed for the chain-conduction optimization mechanism of self-consistency constraints, which provides additional references of solution distributions at a small query cost.
By combining these enhancements, we achieve stable and consistency training in non-pixel modality and latent representation spaces.
Benchmark experiments demonstrate that our method significantly outperforms traditional consistency distillation methods with reduced training cost and enhances the consistency model to perform comparably to state-of-the-art models with lower inference costs.
\end{abstract}


\section{Introduction}
Synthesizing human motion sequences from textual prompts form a robust foundation for multimodal human-computer interactions, including augmented reality and robotic motion imitation \cite{Hu2024,7807322,Shi2024}.
Recent advances \cite{Tevet2023, Zhang2022} have explored a text-driven motion diffusion framework that iteratively solves reverse diffusion trajectories to match motion representation distributions.
These works \cite{Chen2023, Kong2023, jin2024act, Lu2022} demonstrate powerful generative performance at the cost of a hundred-fold increase in computational burden under the iterative solving framework.
Although previous work \cite{Chen2023, Kong2023} introduced higher-order numerical solvers \cite{Liu2022} to solve rapidly in low-dimensional and well-designed latent spaces.
However, larger sampling strides are associated with significant numerical errors due to the nonlinear nature of the diffusion trajectories, causing significantly reduced fidelity of these methods at a lower number of function evaluations (NFE).
Efficiency bottlenecks in motion diffusion frameworks are emerging as a critical bottleneck in their application.

Recent advancements aim to shift computationally expensive iterations to the training phase, enabling large-scale skip-step or single-step sampling during inference by leveraging precomputed diffusion trajectories, known as the \textit{consistency model} \cite{Song2023}.
Typical approaches to precomputed trajectory methods include consistency distillation \cite{Luo2023} and consistency training \cite{Yang2024}.
Consistency distillation relies on a well-trained diffusion model as the teacher, and training it from scratch is resource-intensive and time-consuming.
Additionally, the distillation process is constrained by the sample quality of the teacher model, which caps the performance ceiling.
Conversely, consistency training with lower training costs, which calculates the log probability gradient directly from the raw data, avoids these limitations. 
However, estimating trajectory distributions from individual complex motion samples without distillation guidance is imprecise.
Three key challenges persist:
\textbf{\textit{Firstly}}, human motion sequences are typically represented as dynamic-length and heterogeneous representations due to habitual preferences and motion properties.
Predicting the target distribution from such complex and unbounded continuous solution space with less iterative refinement in the inference phase requires high model robustness.
\textbf{\textit{Secondly}}, vanilla consistency training struggles to fit existing diffusion control techniques and thus works suboptimally in conditional generation scenarios.
Typical examples are conditional trajectory guidance such as Classifier-Free Guidance (CFG) \cite{Ho2021}, which relies heavily on well-trained diffusion models as a prerequisite.
\textbf{\textit{Thirdly}}, the consistency model learns precomputed trajectories through self-consistency constraints. 
It is characterized by a chain conduction mechanism, where the solution distribution for the current perturbed state is determined solely by the preceding perturbed state, leading to inefficient training.
These challenges limit the adaptability of consistency training in complex conditional generation tasks.

\begin{figure*}[!t]
\centerline{\includegraphics[width=\columnwidth]{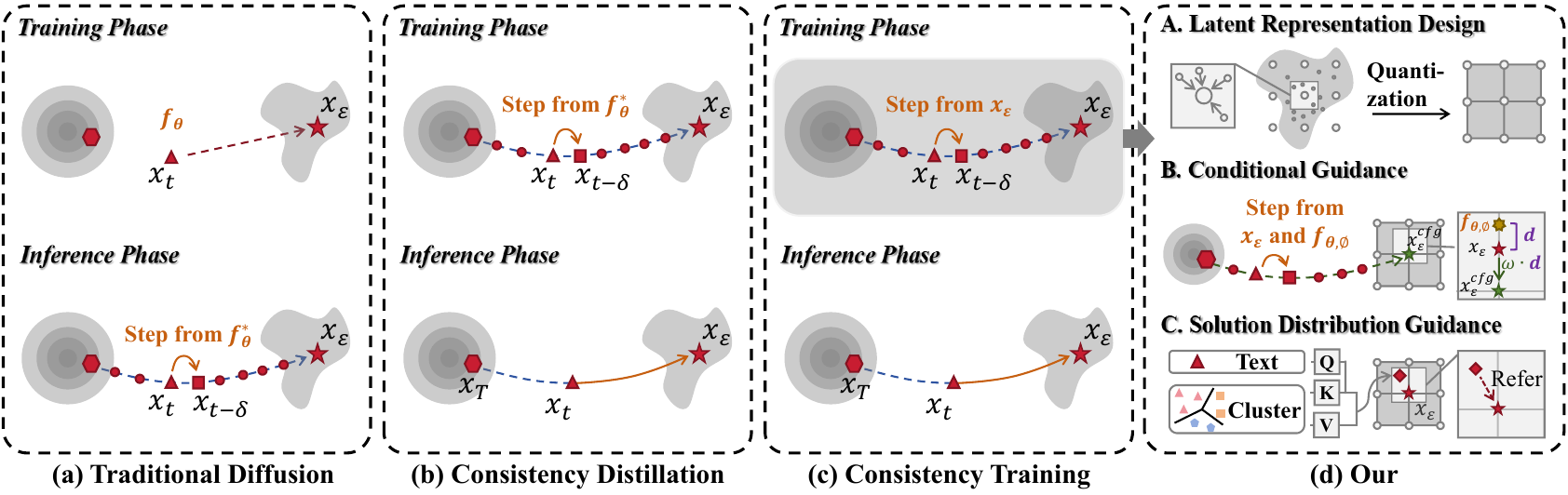}}
\caption{\textbf{Overview of the distinctions between our method and traditional methods.} 
(a) Traditional diffusion methods calculate diffusion trajectories using the well-trained diffusion model $f_\theta^*$ during inference, incurring high sampling iteration costs.
(b) Consistency distillation precomputes the diffusion trajectories in training via the teacher model $f_\theta^*$ and skip-step sampling in inference via self-consistency constraints.
(c) Consistency training eliminates reliance on the teacher model $f_\theta^*$ and estimates diffusion trajectories directly from raw data $x_\epsilon$.
(d) Our approach extends consistency training to the motion latent space by refining motion representations into bounded and concise distributions, integrating conditional guidance to optimize diffusion trajectories from raw data $x_\epsilon$ via an online-trained unconditional model $f_{\theta,\emptyset}$, and introducing a clustering guidance module to supply solution distribution references for given instruction.
}
\label{fig1}
\end{figure*}
To explore the real-time motion diffusion framework with state-of-the-art performance, this paper proposes the motion latent consistency training paradigm from three aspects:


\textit{Latent representation design:} 
Experience with pixel modalities suggests that continuous representations with bounded and finite states simplify solution spaces while enforcing intervals between adjacent states mitigates cumulative errors in precomputed trajectories.
For motion modalities, \uline{our first insight is to extend a motion autoencoder with the quantization constraint to construct pix-like latent representations.}
To this end, we restrict representation boundaries using the hyperbolic tangent (Tanh) function and map continuous representations to their nearest predefined clustering centers.
The significant distinction between traditional variational latent representations and unbounded continuous states lies in our approach, which encodes representations as bounded and finite states.
It avoids the uncertainty variance due to Kullback-Leibler constraints, and the solution space is more concise.
Additionally, previous practice demonstrates that bounded representations contribute to maintaining stable inference in CFG techniques.

\textit{Conditional guidance:} 
Conditional guidance enables controlled enhancement by modulating the gradient difference between the conditional and unconditional probabilities of the current perturbed state.
\uline{Our second insight is to present a conditionally guided consistency training framework through online simulation in the CFG format.}
This framework treats the ground truth latent representation as a simulation of the conditional prediction, replacing the unconditional estimation with an online-updated model driven by an additional loss term.
By eliminating reliance on pre-trained diffusion models, this approach extends the traditional CFG technique beyond the inference phase to the training phase. 
The CFG-based structure further facilitates the differentiation of diffusion trajectories under varying conditions.

\textit{Solution distribution guidance:} 
Constructing retrieval algorithms that leverage cross-modal associations between text and motion representations enables the integration of distribution references into model design.
\uline{Our third insight is to propose an clustering guidance module based on the K-Nearest Neighbor (KNN) algorithm across the entire train set.}
Specifically, KNN is employed to construct clustering dictionaries, where textual representation cluster centers serve as keys, and the mean motion representation values within the same category act as the corresponding values.
It offers additional solution distribution references  beyond the previous state during inference by calculating the correlation between the current textual representation and the clustering dictionary keys to activate the relevant motion categories.

The main contributions of this research are as follows.
\begin{enumerate}
    \item A novel motion latent representation is extended based on the quantization constraint, which is bounded finite states, providing a powerful latent space embedding scheme in the consistency training framework. 

    \item The conditionally guided consistency training framework is established, extending CFG from the inference phase to the training phase. To the best of our knowledge, this work explores consistency training in latent space \textit{for the first time}, and is also \textit{the first} to introduce CFG into consistency training. 

    \item A clustering guidance module based on the KNN algorithm is developed to offer additional solution distribution references with minimal query cost. 

    \item This work achieves performance matching state-of-the-art methods on two datasets: KIT and HumanML, with an inference speed of only 54 ms and without any diffusion model pre-training cost. Extensive experiments indicate the effectiveness of the proposed methods and each component.
    Our code and pre-training weights are available at public repository\footnote{https://github.com/Humengxian/Efficient-Text-driven-Motion-Generation-via-Latent-Consistency-Training}.
\end{enumerate}

\section{Related Works}
\subsection{Human motion generation.} 
Human motion generation focuses on synthesizing motion sequences under specified conditions, such as action categories \cite{Lee2023, Xu2023}, audio inputs \cite{Li2022}, or textual descriptions \cite{Ahuja2019, Tevet2023, Chen2023}.
In recent years, extensive research  \cite{Yan2019, Guo2020} has explored motion generation using diverse generative frameworks.
For instance, VAE-based models \cite{Petrovich2021, Guo2022} encode motion as Gaussian distributions and enforce regularity through KL divergence constraints.
Such constraint allows it to reconstruct the motion information from the standard normal distribution, yet its results are often ambiguous.
GAN-based methods \cite{Wang2020, Cai2018} improve performance by circumventing direct probabilistic likelihood estimation through adversarial training strategies.
However, the adversarial nature of GANs often leads to unstable training and susceptibility to mode collapse.
More recently, multi-step generative approaches, including auto-regressive \cite{Zhang2023, Guo2023} and diffusion methods \cite{Zhang2022, Tevet2023, Chen2023}, have achieved significant success.
Notably, diffusion-based methods are increasingly dominating research frontiers due to their stable distribution estimation capabilities and high-quality sampling results.
MotionDiffuse \cite{Zhang2022} and MDM \cite{Tevet2023} were among the first to adopt diffusion frameworks for motion generation.
Subsequent advancements, such as MLD \cite{Chen2023} and M2DM \cite{Kong2023}, introduced latent space diffusion within variational and discrete representations, respectively, demonstrating that the optimized latent space contributes to improving generation quality and efficiency.
GraphMotion \cite{jin2024act} incorporates semantic role processing tools for fine-grained controllable generation.
ReMoDiffuse \cite{Zhang2023a} further explores hybrid retrieval from train data, enabling the generation of more realistic motion.
These contributions collectively underscore the exceptional capabilities of motion diffusion frameworks.

\subsection{Efficient diffusion sampling.} 
The inefficient diffusion sampling mechanism is the critical bottleneck restricting diffusion models in real-time applications.
The score-based model \cite{Song2021a} established a foundational connection between diffusion frameworks and stochastic differential equations, highlighting a particular case known as the Probability Flow Ordinary Differential Equation (PF-ODE).
It is a milestone achievement. 
Subsequent work based on this theoretical cornerstone is mainly divided into two categories.
The first category is large stride numerical sampling with higher-order ODE approximation solving strategies \cite{Lu2022}.
These solvers serve as plug-and-play generalized diffusion solving strategies, with extensions implemented in previous motion diffusion frameworks such as MLD \cite{Chen2023} and GraphMotion \cite{jin2024act}.
However, numerical solver-based methods are constrained by the nonlinear nature of diffusion trajectories. Recent advancements still require more than 20 function evaluations (NFE) to control numerical errors resulting from large sampling strides.
As a result, these methods struggle to address the efficiency bottleneck of iterative sampling mechanisms.
The second category involves large-scale skip-step sampling aimed at learning diffusion trajectories, commonly called diffusion distillation \cite{Liu2022, Xu2022, Song2023}.
Traditional diffusion distillation methods typically employ a well-trained diffusion model as a teacher to generate precomputed diffusion trajectories.
This approach shifts the complex iterative computation from the inference to the training phase, enabling single-step or few-step generation and alleviating the efficiency bottleneck.
However, distillation-based methods incur higher training costs and face performance limitations.
The latest advances are consistency models \cite{Song2023}, especially the consistency training free from the teacher distillation mode, which has the potential to achieve single-step high-quality generation.

\subsection{Consistency model.} 
Consistency models \cite{Song2023} can be divided into two categories: consistency distillation and consistency training, based on precomputed trajectory methods.
Consistency distillation achieves efficient trajectory distillation and single-step inference by maintaining the consistency of model outputs along the same diffusion trajectory.
This approach relies on a robust teacher model for guidance and is integrated with well-established diffusion model enhancement techniques such as Classifier-Free Guidance (CFG), Lora, and ControlNet.
Stable diffusion guidance further enables its extension to various fields \cite{Luo2023, Kim2023, Ye2023}.
Specifically, we note contemporaneous work that extends consistency distillation to human motion generation tasks.
However, constrained by the high training cost and performance ceiling of the teacher distillation mode, the existing methods remain significant gaps with the state-of-the-art diffusion frameworks. 
Comparatively, consistency training simulates diffusion trajectories within the raw data, freeing from the limitations of pre-trained models.
Nevertheless, its performance is significantly inferior to distillation-based methods due to the lack of guidance, and the advances are stuck in the earliest proposed raw pixel representations.
ICM \cite{Yang2024} further explores and refines consistency training methods to achieve performance comparable to consistency distillation without the need for pre-trained models.
However, its focus remains on raw pixel representations, and the application of consistency training to non-image modalities and latent representations has yet to be explored.
Additionally, previous studies have overlooked guidance techniques in consistency training, such as conditional and solution distribution guidance, which hold significant potential for improvement.
To address these shortcomings, our work focuses on constructing the latent consistency training paradigm to improve the performance of consistency models in motion modalities to state-of-the-art levels with lower training and inference costs.

\begin{figure*}[!t]
\centerline{\includegraphics[width=\columnwidth]{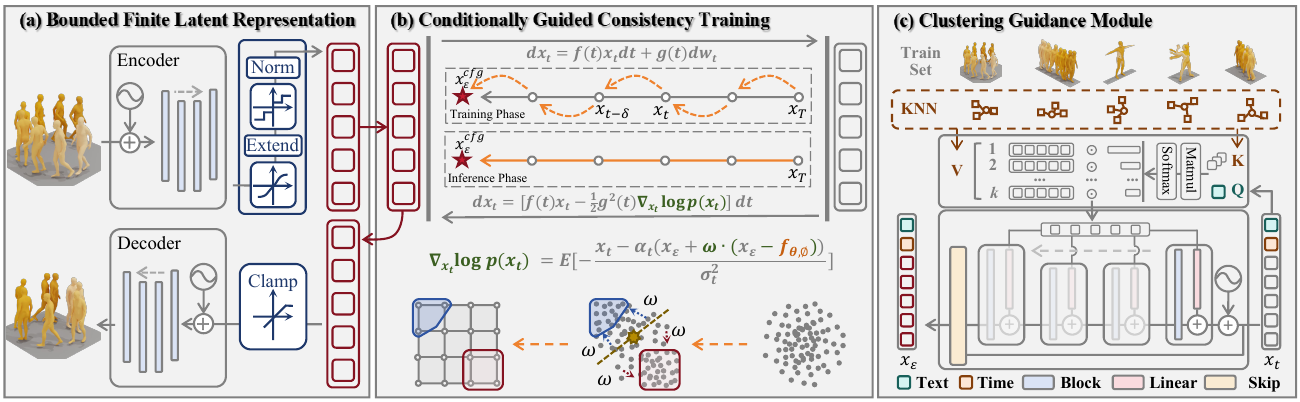}}
\caption{\textbf{Approach overview.} 
(a) Motion sequences are encoded with quantization constraints, ensuring bounded finite states that are structurally analogous to pixel representations.
(b) Conditional diffusion trajectories are constructed during the training phase using an online simulation of the CFG format.
(c) A clustering guidance module is integrated into the consistency model $\mathcal{S}_\psi$. This module constructs a clustering dictionary using the KNN algorithm and leverages an attention-like query mechanism to provide solution distribution references tailored to the given textual conditions.
}
\label{fig2}
\end{figure*}
\section{Preliminaries}
\subsection{Score-based Diffusion Models}
The diffusion model \cite{Ho2020, 10420468} is a generative model that progressively adds Gaussian noise to data and subsequently generates samples from the noise through a reverse denoising process.
Specifically, it transforms the data distribution $p_{data}(x_\epsilon)$ into a well-sampled prior distribution $p(x_T)$ using a Gaussian perturbation kernel $p(x_t|x_\epsilon)=\mathcal{N}(x_t|\alpha_t x_\epsilon,\sigma_t^2I)$, where $\alpha_t$ and $\sigma_t$ are noise schedules.
Recent studies have formalized this process into a continuous-time framework,
\begin{equation}\label{equ1}
    d x_t = f(t)x_tdt + g(t) dw_t,
\end{equation}
where $t\in [\epsilon, T]$, $\epsilon$ and $T$ are the fixed positive constant, $w_{t}$ denotes the standard Brownian motion, $f$ and $g$ are the drift and diffusion coefficients respectively. They are related to the noise schedules as follows.
\begin{equation}\label{equ2}
    f(t) = \frac{d \log \alpha_t}{dt},\quad g^2(t)=\frac{d\sigma_t^2}{dt} - 2\frac{d\log\alpha_t}{dt}\sigma_t^2.
\end{equation}
Previous work has revealed that the reverse process of Equation \ref{equ1} shares the same marginal probabilities with the \textit{probabilistic flow ODE}:
\begin{equation}\label{equ3}
    dx_t = [f(t)x_t - \frac{1}{2}g^2(t)\nabla_{x_t} \log p(x_t)]dt,
\end{equation}
where $\nabla_x \log p(x_t)$ is named the \textit{score function}, which is the only unknown term in the sampling pipeline. 
An effective approach involves training a time-dependent score network, $f_\theta(x_t,t)$ to estimate $\nabla_x \log p(x_t)$ using conditional score matching, which is typically parameterized as the prediction of either the noise or the initial value in the forward diffusion process.
Further, Equation \ref{equ3} can be solved in finite steps by numerical ODE solvers such as Euler \cite{Song2021a} and Heun solvers \cite{Karras2022}.
Upon the above study, recent works have further explored conditional probabilities $p(x_t|y)$ to achieve more controlled and flexible generation, where $y$ represents a given condition, such as textual descriptions or action labels.
One effective approach for controlled generation is Classifier-Free Guidance (CFG), which is parameterized as a linear combination of unconditional and conditional noise, 
\begin{equation}
    \widetilde{f}_\theta(x_t, t, c) =(1 + \omega) f_\theta (x_t, t, c) - \omega f_\theta (x_t, t, \emptyset),
\end{equation}
where $\omega$ is guidance scale.

\subsection{Consistency Models}
Theoretically, the reverse process expressed by Equation \ref{equ3} is deterministic. 
The consistency model \cite{Song2023} achieves one-step or few-step generation by pulling in outputs on the same ODE trajectory.
Given any perturbed state $x_t$ and its previous state $x_{t-\delta}$ on a diffusion trajectory $\{x_T, x_{T-\delta}, \cdots, x_{\epsilon+\delta}, x_\epsilon\}$, the consistency model is formally expressed as,
\begin{equation}\label{equ4}
    \begin{cases} 
 \mathcal{S}_\psi(x_t,t) \approx \mathcal{S}_\psi(x_{t-\delta},t-\delta), \quad \forall t \in [\epsilon+\delta, T], \\
\mathcal{S}_\psi(x_{\epsilon},\epsilon)\approx \epsilon.
\end{cases}
\end{equation}
Here, $\delta$ represents the sampling stride, and $[\epsilon, T]$ denotes the time horizon.
The Equation \ref{equ4} is known as the \textit{\textbf{self-consistency constraints}}.
To maintain the boundary conditions, existing consistency models are commonly parameterized by skip connections, i.e.,
\begin{equation}\label{equ5}
    \mathcal{S}_\psi (x_t, t) := c_{skip} (t) x_t + c_{out}(t) \hat{\mathcal{S}}_\psi (x_t, t),
\end{equation}
where $c_{skip}(t)$ and $c_{out}(t)$ are differentiable functions satisfied $c_{skip}(\epsilon) =1$ and $c_{out}(\epsilon)=0$.
For stabilize training, the consistency model maintaining target model $\mathcal{S}_\psi^-$, trained with the exponential moving average (EMA) of rate $\gamma$: $\psi^- \leftarrow \gamma \psi^- + (1-\gamma)\psi$. 
The consistency loss can be formulated as,
\begin{equation}\label{equ6}
    \mathcal{L}_{cm} = \mathbb{E}_{x,t}\big[d\big(\mathcal{S}_\psi(x_{t},t),\mathcal{S}_{\psi^-}(\hat{x}_{t-\delta},t-\delta)\big)\big]
\end{equation}
where $d(\cdot,\cdot)$ is a metric function. 
The previous state, $\hat{x}{t-\delta}$, represents a one-step estimation of $x_t$ obtained using ODE solvers applied in Equation \ref{equ3}. This approach can be categorized into consistency distillation and consistency training, which estimate the probability gradient $\nabla{x_t} \log p(x_t)$ from the pre-trained diffusion model and raw data, respectively.

As indicated in Equation \ref{equ6}, the output of the current state in the traditional consistency model is exclusively determined by the output of the preceding state, referred to as \textit{\textbf{chain-conduction optimization}}. 
It leads to cumulative errors between the outputs of adjacent perturbed states being transferred to the initial state along the diffusion trajectory and lacks immediate reference to the solution distribution.

\section{Method}
An overview of the proposed method is shown in Figure \ref{fig2}, with Pseudo-codes \ref{train} and \ref{infer} included to clarify the algorithmic workflow and implementation details.
\begin{algorithm*}[tb]
\small
\DontPrintSemicolon
   \caption{Motion Latent Consistency Training.}
   \label{train}
   \KwIn{Train set $\Gamma=\{(x^{(n)}, c^{(n)})\}^N_{n=1}$, Motion AutoEncoder $\mathcal{G}=\{\mathcal{E}, \mathcal{D}\}$ with initial parameter $\theta$, size $2l+1$ of finite set $\mathcal{M}$, Joint Transform Function $\mathcal{J}$, Motion Consistency Model $\mathcal{S}$ with initial parameter $\psi$ and $\psi^-$, ODE Solver $\Phi$, Timestep Scheduler $\{t_i\}_{i=0}^I$, Guidance Scale $\omega$, Learning Ratio $\eta$, EMA Ratio $\gamma$, Balance Weight $\lambda_j$;}
   
   \textbf{\# Stage 1: Motion AutoEncoder Training.}

    \Repeat{convergence}{
    Sample motion $x \sim \Gamma$;

    $z_e \leftarrow \mathcal{E}(x)$;    \tcp*[r]{Motion Encoding.}

    $z_m \leftarrow\mathcal{R}\Big(l \cdot tanh(z_e)\Big)/l$; \tcp*[r]{Quantization Constraint.} 

    $\mathcal{L}_{z}\leftarrow\mathbb{E}_x \Big[d\Big(x, \mathcal{D}(z_m)\Big)+ \lambda_j d\Big(\mathcal{J}(x), \mathcal{J}(\mathcal{D}(z_m))\Big)\Big]$;\tcp*[r]{Loss.}

    $\theta \leftarrow \theta - \eta \nabla_\theta \mathcal{L}_{z}$. \tcp*[r]{Update $\theta$.}
    }

    \textbf{\# Stage 2: Motion Consistency Training.}

    \Repeat{convergence}{
    Sample motion $x$ and condition $ c \sim \Gamma$, noise $z\sim \mathcal{N}(0, I)$, timestep $t_i, t_{i-1}\sim \{t_i\}_{i=0}^I$;
    
    $x_\epsilon \leftarrow\mathcal{R}\Big(l \cdot tanh(\mathcal{E}(x))\Big)/l$; \tcp*[r]{Motion Encoding.}
    
    $x_{t_i} \leftarrow \alpha_{t_i} \cdot x_\epsilon + \sigma_{t_i} \cdot z$; \tcp*[r]{ Perturbed Data. $\alpha_{t}$ and $\sigma_{t}$ Detailed in Equation \ref{equ10}.}
    
    $x_{\epsilon}^{\Phi}\leftarrow  (1+\omega) \cdot x_\epsilon - \omega \mathcal{S}_\psi(x_{t_i}, t_i, \emptyset);$  \tcp*[r]{CFG in Consistency Training.} 

    $x_{\epsilon}^{\Phi} \leftarrow \text{clamp}(x_{\epsilon}^{\Phi}, -1, 1);$  \tcp*[r]{Clamp.}
    
    $\widetilde{x}_{t_{i-1}} \leftarrow \Phi (x_{\epsilon}^{\Phi}, t_i, t_{i-1})$;  \tcp*[r]{One-step Numerical Estimation with Equation \ref{equ12}.}
    
    $\mathcal{L}_{c}\leftarrow\mathbb{E}_{x,t} \Big [ \frac{1}{t_i - t_{i-1}} d\Big(\mathcal{S}_\psi (x_{t_i},t_i,c), \mathcal{S}_{\psi^-}(\widetilde{x}_{t_{i-1}}, t_{i-1}, c)\Big) + d\Big(\mathcal{S}_\psi(x_{t_i}, t_i, \emptyset), x_\epsilon \Big) \Big]$;       \tcp*[r]{Loss.} 
    
    $\psi \leftarrow \psi - \eta \nabla_\psi \mathcal{L}_{c}$; \tcp*[r]{ Update $\psi$.}

    $\psi^-\leftarrow \text{stopgrad}(\gamma \psi^- + (1-\gamma)\psi).$ \tcp*[r]{ Update $\psi^-$.}
    }
\end{algorithm*}
\begin{algorithm*}[tb]
\DontPrintSemicolon
\SetKwFor{KwFor}{for}{\KwTo}{do}
\small
   \caption{Motion Latent Consistency Inferring.}
   \label{infer}
\KwIn{Motion AutoEncoder $\mathcal{G}=\{\mathcal{E}, \mathcal{D}\}$, Motion Consistency Model $\mathcal{S}_\psi$, Condition $c$, Max Number of Function Evaluations $N$, Timestep Scheduler $\{t_i\}_{i=0}^{N-1}$;}
\KwResult{Motion Sequence $x$.}

Sample $x_{t_N},z \sim \mathcal{N}(0, I)$;

\For{i=N-1 $\text{to}$ 1}{
\If{i != 1}{
    $x_{t_i} \leftarrow \alpha_{t_i} \cdot x_\epsilon + \sigma_{t_i} \cdot z$;  \tcp*[r]{Perturbed Data.}
}
    $x_\epsilon^{pred}  \leftarrow \mathcal{S}_\psi (x_{t_i},t_i,c)$; \tcp*[r]{Denoising.}

    $x_\epsilon^{pred}  \leftarrow \text{clamp}(x_\epsilon^{pred}, -1, 1)$; \tcp*[r]{Clamp.}
}
$x = \mathcal{D}(x_t)$. \tcp*[r]{Motion Decoding.}
\end{algorithm*}

\subsection{Bounded Finite Latent Representation}\label{sec4.1}
Raw human motion sequences are typically represented as a heterogeneous combination of joint rotations, positions, velocities, and other related features.
For motion representation optimization, a motion autoencoder $\mathcal{G}=\{\mathcal{E},\mathcal{D}\}$ is constructed for encoding and reconstruction between the raw motion sequence $x$ and the motion latent representation $z$.
It relies on quantization constraints to ensure boundedness and regularity of $z$.
Specifically, each dimension of $z$ is sampled from a finite set $\mathcal{M}$ of size $2l+1$ as follow,
\begin{equation}
    \mathcal{M} = \{z_i;-1, -j/l, \cdots, 0, \cdots, j/l, \cdots, 1\}_{j=0}^l.
\end{equation}

For clarity, let $l$ denote the quantization level, where $j/l$ represents the predefined clustering center. 
It is structurally analogous to normalized pixel representations, characterized by its distinctive features of finite continuous states and enforced intervals between adjacent states.
The motion latent representation $z\in\mathbb{R}^{n,d}$ is defined as $n$ learnable tokens with $d$-dimensional features, effectively aggregating the motion sequence information through attention-based computation \cite{Vaswani2017}. 
The encoder $\mathcal{E}$ applies a hyperbolic tangent (\textit{Tanh}) activation to constrain the outputs within bounded limits, followed by quantization using the rounding operator $\mathcal{R}$.
To enable backpropagation, the quantized values approximate their gradients using the gradient of the unquantized state, known as the straight-through estimator (STE) \cite{Bengio2013}.
The sampled latent representations $z_m$ adhere to the following format:
\begin{equation}
    z_m =\mathcal{R}\Big(l \cdot tanh(\mathcal{E}(x))\Big)/l.
\end{equation}

The standard optimization target is to reconstruct motion information from $z$ with the decoder $\mathcal{D}$, i.e., to optimize the $l_1$ smooth error loss,
\begin{equation}\label{equ9}
\mathcal{L}_{z}=\mathbb{E}_x \Big [ d\Big(x, \mathcal{D}(z_m)\Big)+ \lambda_j d\Big(\mathcal{J}(x), \mathcal{J}(\mathcal{D}(z_m))\Big)\Big  ],
\end{equation}
where $\mathcal{J}$ is a function to transform features such as joint rotations into joint coordinates, $\lambda_j$ is the balancing weight.
\subsection{Conditionally Guided Consistency Training}\label{sec4.2}
The diffusion stage begins with the variance preserving schedule \cite{Song2021a} to perturbed motion latent representations $x_\epsilon=z$ with perturbation kernel  $\mathcal{N}(x_t;\alpha(t) x_\epsilon,\sigma^2(t)I)$,
\begin{equation}\label{equ10}
    \alpha_t := e^{-\frac{1}{4} t^2 (\beta_1 - \beta_0) - \frac{1}{2}t\beta_0}, \quad \sigma_t:=\sqrt{1 - e^{2\alpha(t)}}.
\end{equation}
The consistency model $\mathcal{S}_\theta$ has been constructed to predict $x_\epsilon$ from perturbed $x_t$ in a given PF-ODE trajectory. 
To maintain the boundary conditions that $\mathcal{S}_\psi(x_\epsilon, \epsilon, c) = x_\epsilon$, we employ the same skip setting for Equation \ref{equ5} as in LCM \cite{Luo2023}, which parameterized as follow:

\begin{equation}
\mathcal{S}_\psi(x_t, t, c) := \frac{\eta ^ 2}{(10t)^2+\eta^2} \cdot x_t + \frac{10t}{\sqrt{(10t)^2+\eta ^2}} \cdot \widetilde{\mathcal{S}}_\psi(x_t, t, c),
\end{equation}
where $\widetilde{\mathcal{S}}_\psi$ is a transformer-based network, and $\eta$ is a hyperparameter usually set to 0.5.  
Following \textit{the self-consistency property} (as detailed in Equation \ref{equ4}), the model $\mathcal{S}_\psi$ has to maintain the consistency of the output at the given perturbed state $x_t$ with the previous state $\hat{x}_{t-\delta}$ on the same ODE trajectory.
The latter can be estimated from Equation \ref{equ3} by DPM-Solver++:
\begin{equation}\label{equ12}
    \widetilde{x}_{t-\delta} =  \frac{\sigma_{t-\delta}}{\sigma_t}\cdot x_t - \alpha_{t-\delta} \cdot (e^{-h_t}-1) \cdot x_{\epsilon}^{\Phi},
\end{equation}
where $h_t:=\lambda_{t-\delta} - \lambda_{t}$, $\lambda_t:=\log (\alpha_t/\sigma_t)$, and $x_{\epsilon}^{\Phi}$ is the estimation of $x_\epsilon$ under the different sampling strategies.
In particular,  $x_{\epsilon}^{\Phi}$ can be parameterized as a linear combination of conditional and unconditional latent presentation prediction following the CFG strategy, i.e.,
\begin{equation}\label{equ13}
    x_{\epsilon}^{\Phi}(x_t, t, c) = (1+\omega) \cdot \mathcal{F}_\psi(x_t, t, c) - \omega \mathcal{F}_\psi(x_t, t, \emptyset),
\end{equation}
where $\mathcal{F}_\psi(\cdot)$ is well-trained and $x_\epsilon$-prediction-based motion diffusion model.
Unconditional prediction models estimate the average observation of the current perturbation state without considering conditional differences. When large perturbations, $x_{\epsilon}^{\Phi}$ moves toward the conditional gradient, emphasizing differences from the average state. As perturbations decrease, unconditional predictions align with conditional predictions, achieving convergence.

It is worth noting that $x_\epsilon$ can be utilized to simulate $\mathcal{F}_\psi(x_t, t, c)$ as used in the vanilla consistency training pipeline.
Furthermore, $\mathcal{F}_\psi(x_t, t, \emptyset)$ can be replaced by  $\mathcal{S}_\psi(x_t, t, \emptyset)$  with online updating based on the additional unconditional loss item.
Thus Equation \ref{equ13} can be rewritten as:
\begin{equation}\label{equ14}
    x_{\epsilon}^{\Phi}(x_t, t, c) = (1+\omega) \cdot x_\epsilon - \omega \mathcal{S}_\psi(x_t, t, \emptyset).
\end{equation}
We refer to Equation \ref{equ14} as the \textit{conditional trajectory simulation}. The optimization objective is that,
\begin{equation}
\begin{aligned}
    \mathcal{L}_{c}=\mathbb{E}_{x,t} \Big [ &\underbrace{\frac{1}{\delta} d\Big(\mathcal{S}_\psi (x_t,t,c), \mathcal{S}_{\psi^-}(\widetilde{x}_{t-\delta}, t-\delta, c)\Big)}_{\text{Consistency Loss}} \\
    &+ \underbrace{d\Big(\mathcal{S}_\psi(x_t, t, \emptyset), x_\epsilon \Big)}_{\text{Unconditional Loss}} \Big],
\end{aligned}
\end{equation}
where $d(x,y)$ is pseudo-huber metric. 
The target network $\mathcal{S}_{\psi^-}$  is updated after each iteration via EMA.

\subsection{Clustering Guidance Module} \label{sec4.3}
For the chain-conduction optimization mechanism with self-consistency constraints, the clustering guidance module is proposed in the consistency model design to enhance the solution distribution guidance under specific textual conditions. 

Before training, a clustering dictionary is constructed from the training set.
Specifically, the K-Nearest Neighbor (KNN) algorithm is applied to classify the embedded features of each text in the training set into $k$ clusters.
The clustering centers for each class are utilized as keys to construct the clustering dictionary, denoted as $\mathcal{K}\in \mathbb{R}^{K,d_c}$, where $d_c$ represents the dimension of the text representations. 
Subsequently, we calculate the mean values of the corresponding motion representations within the same text categories to construct the values of the clustering dictionary, denoted as $\mathcal{V}\in \mathbb{R}^{K,n,d_m}$, where $n$ is the token count, and $d_m$ is the dimension of the motion representations.

In the training and inference phases, the clustering dictionary is invoked via the similarity query mechanism.
For instance, given an instruction as a query vector, denoted as $\mathcal{Q}\in \mathbb{R}^{1,d_c}$.
$\mathcal{Q}$ and $\mathcal{K}$ are projected to the higher dimensional space via affine mapping $\mathcal{A}$, then the similarity is computed,
\begin{equation}
    \rho = softmax(\mathcal{\mathcal{A}(Q)}\cdot \mathcal{A}(K)^T).
\end{equation}

The clustering representation $\mathcal{I}\in \mathbb{R}^{1,n,d_m}$ computed as:
\begin{equation}
    \mathcal{I} = \rho \cdot \mathcal{V}.
\end{equation}

The clustering guidance offers a more flexible scheme that allows the model to localize the solution distribution at a lower query cost rapidly.
The query computation is performed only once during a single inference process to manage computational complexity. 
For the input $x^{(i)}$ of the $i$-th block in the backbone network, we map the query results $\mathcal{I}$ into dimensions consistent with the $x^{(i)}$ through affine mapping and implement feature fusion using an element-wise summation operator.

\section{Experiments}
\subsection{Experimental Setup}\label{sec5.1}
\subsubsection{Datasets} We evaluate our framework on two widely used benchmarks for text-driven motion generation: KIT \cite{Plappert2016} and HumanML3D \cite{Guo2022}.
The KIT dataset comprises 3,911 motion sequences paired with 6,363 natural language descriptions.
The HumanML3D dataset, a combination of HumanAct12 \cite{Guo2020} and AMASS \cite{Mahmood2019}, contains 14,616 motion sequences and 44,970 textual descriptions.

\begin{table*}[t]
\setlength{\tabcolsep}{3pt}
\begin{center}
\caption{Comparisons to state-of-the-art methods on the HumanML3D test set. "$\uparrow$" denotes that higher is better. "$\downarrow$" denotes that lower is better. "$\rightarrow$" denotes that results are better if the metric is closer to the real motion. The gray background indicates the sota method of the current framework.
\textbf{Bold} and \underline{underlined} indicate the best and second-best results, respectively.
AITS is first measured on the RTX 4090 GPU and then aligned on the Tesla V100 GPU using the MLD as an intermediary benchmark.
}
\small
    \begin{tabular}{lC{1.7cm}C{1.7cm}C{1.7cm}C{1.7cm}C{1.7cm}C{1.7cm}C{1.7cm}C{1.7cm}}
    \toprule
    \multirow{2}[4]{*}{Method} & \multirow{2}[4]{*}{AITS (s) $\downarrow$}   & \multicolumn{3}{c}{R-Precision $\uparrow$} & \multirow{2}[4]{*}{FID $\downarrow$} & \multirow{2}[4]{*}{MM-Dist $\downarrow$} & \multirow{2}[4]{*}{Diversity $\rightarrow$} & \multirow{2}[4]{*}{MModality $\uparrow$} \\
\cmidrule{3-5}   &   & Top-1 & Top-2 & Top-3 &       &       &       &         \\
    \midrule
    Real & - & $0.511^{\pm .003}$ & $0.703^{\pm .003}$ & $0.797^{\pm .002}$ & $0.002^{\pm .000}$ & $2.974^{\pm .008}$ & $9.503^{\pm .065}$ & -     \\
    \midrule
    MDM \cite{Tevet2023} & 24.74 & $0.320^{\pm .005}$ & $0.498^{\pm .004}$ & $0.611^{\pm .007}$ & $0.544^{\pm .044}$ & $5.566^{\pm .027}$ & $\underline{9.559}^{\pm .086}$ & $\textbf{2.799}^{\pm .072}$ \\
    MLD \cite{Chen2023} & 0.217  & $0.481^{\pm .003}$ & $0.673^{\pm .003}$ & $0.772^{\pm .002}$ & $0.473^{\pm .013}$ & $3.196^{\pm .010}$ & $9.724^{\pm .082}$ & $2.413^{\pm .079}$ \\
    GraphMotion \cite{jin2024act} & 1.495 & $0.504^{\pm .003}$ & $0.699^{\pm .002}$ & $0.785^{\pm .002}$ & $0.116^{\pm .007}$ & $3.070^{\pm .008}$ & $9.692^{\pm .067}$ & $\underline{2.766}^{\pm .096}$  \\
   \rowcolor{gray!30} ReMoDiffuse \cite{Zhang2023a} & 0.417  & $0.510^{\pm .005}$ &  $0.698^{\pm .006}$ &  $0.795^{\pm .004}$ &  $0.103^{\pm .004}$ &  $2.974^{\pm .016}$ &  $9.018^{\pm .075}$ &  $1.795^{\pm .043}$   \\
    \midrule
    T2M-GPT \cite{Zhang2023} & 0.598   & $0.491^{\pm .003}$ & $0.680^{\pm .003}$ & $0.775^{\pm .002}$ & $0.116^{\pm .004}$ & $3.118^{\pm .011}$ & $9.761^{\pm .081}$ & $1.856^{\pm .011}$ \\
    AttT2M \cite{Zhong_2023_ICCV} & 0.717  &  $0.499^{\pm .003}$ &  $0.690^{\pm .002}$ &  $0.786^{\pm .002}$ &  $0.112^{\pm .006}$ &  $3.038^{\pm .007}$ &  $9.700^{\pm .090}$ &  $2.452^{\pm .051}$\\
   \rowcolor{gray!30} MoMask \cite{Guo2023} & 0.118  & $0.521^{\pm .002}$ &  $0.713^{\pm .002}$ &  $0.807^{\pm .002}$ & $\textbf{0.045}^{\pm .002}$ &  $2.958^{\pm .008}$ & - &  $1.241^{\pm .040}$   \\
    \midrule
    Our (NFE 1)& \textbf{0.031} & $0.530^{\pm .002}$ & $0.726^{\pm .002}$ & $0.822^{\pm .002}$ & $0.264^{\pm .007}$ & $2.888^{\pm .007}$ & $9.799^{\pm .061}$ & $2.188^{\pm .049}$  \\
    Our (NFE 2) & \underline{0.038} & $\textbf{0.538}^{\pm .003}$ & $\textbf{0.734}^{\pm .002}$ & $\textbf{0.828}^{\pm .002}$ & $0.094^{\pm .003}$ & $\underline{2.822}^{\pm .005}$ & $9.595^{\pm .075}$ & $2.325^{\pm .061}$ \\
   \rowcolor{gray!30} Our (NFE 4) & 0.054 & $\underline{0.537}^{\pm .003}$ & $\underline{0.732}^{\pm .002}$ & $\underline{0.826}^{\pm .002}$ & $\underline{0.060}^{\pm .003}$ & $\textbf{2.819}^{\pm .010}$ & $\textbf{9.545}^{\pm .068}$ &  $2.571^{\pm .051}$ \\ 
    \bottomrule
    \end{tabular}%
  \label{tab1}%
\end{center}
\end{table*}

\begin{table*}[t]
\begin{center}
\setlength{\tabcolsep}{3pt}
\caption{Comparisons to state-of-the-art methods on the KIT test set. The meaning of the marks and symbols used in this comparison follows the same conventions as described in Table \ref{tab1}.}
\small
    \begin{tabular}{lC{1.7cm}C{1.7cm}C{1.7cm}C{1.7cm}C{1.7cm}C{1.7cm}C{1.7cm}C{1.7cm}}
    \toprule
    \multirow{2}[4]{*}{Method} & \multirow{2}[4]{*}{NFE $\downarrow$}  & \multicolumn{3}{c}{R-Precision $\uparrow$} & \multirow{2}[4]{*}{FID $\downarrow$} & \multirow{2}[4]{*}{MM-Dist $\downarrow$} & \multirow{2}[4]{*}{Diversity $\rightarrow$} & \multirow{2}[4]{*}{MModality $\uparrow$} \\
\cmidrule{3-5} &  & Top-1 & Top-2 & Top-3 &       &       &       &         \\
    \midrule
    Real & -  & $0.424^{\pm .005}$ & $0.649^{\pm .006}$ & $0.779^{\pm .006}$ & $0.031^{\pm .004}$ & $2.788^{\pm .012}$ & $11.08^{\pm .097}$ & -     \\
    \midrule
    MDM \cite{Tevet2023} & 1000 & $0.164^{\pm .004}$ & $0.291^{\pm .004}$ & $0.396^{\pm .004}$ & $0.497^{\pm .021}$ & $9.191^{\pm .022}$ & $10.85^{\pm .109}$ & $1.907^{\pm .214}$  \\
    MLD \cite{Chen2023} & 50  & $0.390^{\pm .008}$ & $0.609^{\pm .008}$ & $0.734^{\pm .007}$ & $0.404^{\pm .027}$ & $3.204^{\pm .027}$ & $10.80^{\pm .117}$ & $2.192^{\pm .071}$  \\
     GraphMotion \cite{jin2024act} & 150 & $0.429^{\pm .007}$ & $0.648^{\pm .006}$ & $0.769^{\pm .006}$ & $0.313^{\pm .013}$ & $3.076^{\pm .022}$ & $\textbf{11.12}^{\pm .135}$ & $\textbf{3.627}^{\pm .113}$  \\
  \rowcolor{gray!30}  ReMoDiffuse  \cite{Zhang2023a} & 50  & $0.427^{\pm .014}$ & $0.641^{\pm .004}$ & $0.765^{\pm .055}$ & $\underline{0.155}^{\pm .006}$ & $2.814^{\pm .012}$ & $10.80^{\pm .105}$ & $1.239^{\pm .028}$  \\
    \midrule
     T2M-GPT \cite{Zhang2023} & 51 & $0.416^{\pm .006}$ & $0.627^{\pm .006}$ & $0.745^{\pm .006}$ & $0.514^{\pm .029}$ & $3.007^{\pm .023}$ & $10.921^{\pm .108}$ & $1.570^{\pm .0.39}$  \\
     AttT2M \cite{Zhong_2023_ICCV} & - & $0.413^{\pm .006}$ &  $0.632^{\pm .006}$ &  $0.751^{\pm .006}$ & $0.870^{\pm .039}$ &  $3.039^{\pm .021}$ & $10.96^{\pm .043}$ &  $\underline{2.281}^{\pm .047}$   \\
   \rowcolor{gray!30}  MoMask \cite{Guo2023} & - & $0.433^{\pm .007}$ &  $0.656^{\pm .005}$ &  $0.781^{\pm .005}$ & $0.204^{\pm .011}$ &  $\textbf{2.779}^{\pm .022}$ & - &  $1.131^{\pm .043}$   \\
    \midrule
     Our (NFE 1) & \textbf{1}  & $ 0.438^{\pm .008}$ & $\underline{0.662}^{\pm .007}$ & $0.785^{\pm .007}$ & $0.253^{\pm .012}$ & $2.822^{\pm .022}$ & $11.138^{\pm .094}$ & $1.718^{\pm .030}$ \\
     Our (NFE 2)& \underline{2}  & $ \textbf{0.443}^{\pm .007}$ & $\textbf{0.670}^{\pm .008}$ & $\textbf{0.795}^{\pm .007}$ & $0.186^{\pm .006}$ & $\underline{2.783}^{\pm .021}$ & $11.137^{\pm .107}$ & $1.550^{\pm .029}$ \\
   \rowcolor{gray!30} Our (NFE 4) & 4 & $ \underline{0.440}^{\pm .007}$ & $\underline{0.662}^{\pm .007}$ & $\underline{0.789}^{\pm .006}$ & $ \textbf{0.153}^{\pm .005}$ & $2.813^{\pm .019}$ & $\underline{11.024}^{\pm .094}$ & $1.657^{\pm .034}$  \\
    \bottomrule
    \end{tabular}
  \label{tab2}%
\end{center}
\end{table*}

\begin{table*}[t]
\begin{center}
\caption{Ablation study about each part of our method on the HumanML3D test set. $\checkmark$ and \ding{55} denote enabled and disabled, respectively.}
\small
    \begin{tabular}{c@{\hspace{14pt}}c@{\hspace{14pt}}c|@{\hspace{14pt}}c@{\hspace{12pt}}c@{\hspace{12pt}}c@{\hspace{12pt}}c@{\hspace{12pt}}c}
    \toprule
   \makecell[c]{Representation \\ Constraint} & \makecell[c]{Conditionally \\ Guided CT} &  \makecell[c]{Clustering \\ Guidance} & \makecell[c]{R-Precision \\ Top-3 $\uparrow$} & FID $\downarrow$ & MM-Dist $\downarrow$ & \makecell[c]{Diversity $\rightarrow$ \\ (9.503)} & MModality $\uparrow$ \\
    \midrule
  None & \ding{55} & \ding{55}  & $0.602^{\pm .005}$ & $2.837^{\pm .028}$ & $5.311^{\pm .127}$ & $8.239^{\pm .035}$ &  $4.113^{\pm .072}$ \\
    \midrule
  Kullback-Leibler & \ding{55} & \ding{55}  & $0.639^{\pm .006}$ & $2.651^{\pm .021}$ & $4.021^{\pm .103}$ & $8.421^{\pm .040}$ &  $3.909^{\pm .040}$ \\
  Kullback-Leibler & \checkmark & \ding{55}  & $0.778^{\pm .005}$ & $0.541^{\pm .008}$ & $3.201^{\pm .008}$ & $9.012^{\pm .093}$ &  $2.570^{\pm .042}$ \\    
  Kullback-Leibler & \ding{55} & \checkmark  & $0.634^{\pm .004}$ & $2.596^{\pm .010}$ & $4.036^{\pm .007}$ & $9.401^{\pm .086}$ &  $4.063^{\pm .065}$ \\   
  Kullback-Leibler & \checkmark & \checkmark  & $0.789^{\pm .003}$ & $0.326^{\pm .006}$ & $2.990^{\pm .007}$ & $9.434^{\pm .027}$ &  $2.417^{\pm .068}$ \\  
    \midrule
  Quantization & \ding{55} & \ding{55}  & $0.734^{\pm .004}$ & $0.615^{\pm .006}$ & $3.351^{\pm .008}$ & $9.248^{\pm .084}$ &  $3.961^{\pm .059}$ \\   
  Quantization & \checkmark & \ding{55}  & $0.821^{\pm .002}$ & $0.210^{\pm .005}$ & $2.886^{\pm .009}$ & $9.535^{\pm .069}$ &  $2.411^{\pm .050}$ \\   
 Quantization & \ding{55} & \checkmark  & $0.733^{\pm .004}$ & $0.542^{\pm .005}$ & $3.351^{\pm .007}$ & $9.234^{\pm .055}$ &  $4.079^{\pm .043}$ \\  
 \midrule
 Quantization & \checkmark & \checkmark  & $0.826^{\pm .002}$ & $0.060^{\pm .003}$ & $2.819^{\pm .010}$ & $9.545^{\pm .068}$ &  $2.571^{\pm .051}$ \\ 
    \bottomrule
    \end{tabular}
  \label{table3}%
\end{center}
\end{table*}

\subsubsection{Evaluation Metrics} 
Aligned with previous studies \cite{Tevet2023}, our evaluation of the proposed framework encompasses four key aspects.
(a) Motion quality: we utilize the \textbf{frechet inception distance (FID)} to evaluate the distance in feature distribution between the generated data and the real data.
(b) Condition matching: we first employ the \textbf{R-precision} to measure the correlation between the text description and the generated motion sequence and record the probability of the first $k=1,2,3$ matches.
Then, we further calculate the distance between motions and texts by \textbf{multi-modal distance (MM Dist)}.
(c) Motion diversity: we compute differences between features with the \textbf{diversity} metric and then measure generative diversity in the same text input using \textbf{multimodality (MM)} metric.
(d) Calculating burden: we first use \textbf{the number of function evaluations (NFE)} to evaluate the resource consumption required for inferring. 
We affirm that the NFE metric is only evaluated in this paper in the context of the reverse diffusion stage, as reported in LCM.
Then, we further utilize \textbf{the Average Inference Time per Sentence (AITS)} measured in seconds to evaluate the inference efficiency of diffusion models. 

\subsubsection{Model Parameters}
Our models are based on transformer architectures with long skip connections \cite{Ronneberger2015}.
Specifically, the encoder $\mathcal{E}$ and decoder $ \mathcal{D}$ contain 7 layers of transformer blocks with input dimensions 256, and each block contains 4 learnable tokens. 
The quantized levels $l$ default setting as 256, and the guidance scale $\omega$ default setting as 4 on the HumanML3D and 5 on the KIT. 
The consistency model $\mathcal{S}$ contains 9 layers of transformer blocks with input dimensions 512.
Following previous work \cite{Dai2024}, we use the frozen \textit{sentence-t5-large} model \cite{ni-etal-2022-sentence} as the text encoder.
The text encoder encodes the text into a pooled output $p\in \mathcal{R}^{1,768}$ and then projects it as text embedding to sum with the time embedding before the input of each block.

\subsubsection{Train Parameters}
For the diffusion strategy \ref{equ10}, we follow the general settings of $\beta_0 = 0.002$ and $\beta_1 = 1$.
For time horizon $[\epsilon, T]$ into $N - 1$ sub-intervals, with $\epsilon$ set to 0.002, $T$ set to 1, $N$ set to 50. 
We follow the consistency model \cite{Song2023} to determine $t_i=(\epsilon^{1/ \rho}+ \frac{i-1}{N-1}(T^{1/\rho}-\epsilon^{1/\rho}))^\rho$, where $\rho=7$.
The EMA rate $\gamma$ is set to 0.995 on the HumanML3D and 0.9 on the KIT.
All experiments except those specifically stated were configured by default to 4 NFE.
For balance training, the weight $\lambda_j$ is set to $10^{-3}$ on the HumanML3D and $10^{-5}$ on the KIT.
All the proposed models are trained with the AdamW optimizer with a learning rate of $10^{-4}$. 

\subsection{Comparisons to State-of-the-art Methods}
The quantitative test results for the HumanML and KIT datasets are provided in Tab. \ref{tab1} and \ref{tab2}, respectively. 
The results are categorized into three areas: previous diffusion frameworks, non-diffusion generative frameworks, and our proposed framework. 
Consistent with prior research \cite{Tevet2023, Chen2023}, we conducted all evaluations 20 times and reported the averages with a 95\% confidence interval. 
Experiments show that the proposed framework achieves state-of-the-art performance at a minimal inference cost.

On the \textit{HumanML3D} dataset, the proposed framework surpasses previous motion diffusion frameworks \cite{Zhang2023a} across multiple metrics, maintaining high diversity while enhancing controllability and reducing inference costs by over 70\%.
Additionally, our single-step inference is competitive, surpassing the baseline of motion latent diffusion models \cite{Chen2023}.
For recent advances \cite{Guo2023} in masked transformer models, our approach matches the FID metrics while reducing inference cost by 50\% and demonstrating superior controllability and diversity.

On the smaller datasets, such as the \textit{KIT} dataset, the proposed framework excels in controllability and FID, achieving solid results with fewer NFEs.
On the KIT dataset, the model’s diversity is reduced compared to HumanML, likely due to the inherent challenges of accurately modeling motion distributions with limited data.
Notably, this trend aligns with observations from prior state-of-the-art generative frameworks, including ReMoDiffuse and MoMask.

\begin{table*}[t] %
    \begin{minipage}{.48\textwidth} 
    \label{table5}
    \small
        \centering  
        \caption{Ablation study of different token counts $n$ on the HumanML3D test set. The gray background indicates the default parameters.}
          \setlength{\tabcolsep}{2.8mm}{
           \label{table4}
        \begin{tabular}{cccc}  
            \toprule  
            $n$  & \makecell[c]{R-Precision\\ Top-3$\uparrow$} & FID $\downarrow$ & MModality $\uparrow$ \\  
            \midrule  
            1 & $0.810^{\pm .002}$  & $0.249^{\pm .009}$  & $2.935^{\pm .067}$  \\  
            2 & $0.804^{\pm .003}$  & $0.210^{\pm .005}$  & $2.872^{\pm .069}$  \\  
            3 & $0.814^{\pm .003}$  & $0.136^{\pm .005}$  & $2.828^{\pm .068}$  \\  
           \rowcolor{gray!30} 4 & $0.826^{\pm .002}$  & $0.060^{\pm .003}$  & $2.571^{\pm .051}$  \\    
            5 & $0.826^{\pm .002}$  & $0.094^{\pm .010}$  & $2.716^{\pm .065}$  \\  
            \bottomrule  
        \end{tabular}  
        }
    \end{minipage}%
    \hspace{0.4cm}
    \begin{minipage}{.48\textwidth} 
    \small
        \centering  
         \caption{Ablation study of different quantization levels $l$ on the HumanML3D test set. The gray background indicates the default parameters.} 
         \setlength{\tabcolsep}{3mm}{
         \label{table5}
        \begin{tabular}{cccc}  
            \toprule  
            $l$ & \makecell[c]{R-Precision\\ Top-3 $\uparrow$} & FID $\downarrow$& MModality $\uparrow$ \\  
            \midrule  
            128 & $0.814^{\pm .002}$  & $0.113^{\pm .004}$  & $2.612^{\pm .065}$  \\
            \rowcolor{gray!30} 256 & $0.826^{\pm .002}$  & $0.060^{\pm .003}$  & $2.571^{\pm .051}$  \\    
            512 & $0.825^{\pm .003}$  & $0.121^{\pm .005}$  & $2.721^{\pm .043}$  \\  
             1024 & $0.812^{\pm .003}$  & $0.142^{\pm .005}$  & $2.872^{\pm .072}$  \\  
            2048 & $0.819^{\pm .002}$  & $0.134^{\pm .007}$  & $2.848^{\pm .066}$  \\    
            \bottomrule 
        \end{tabular} 
        }
    \end{minipage}   
\end{table*}  

\subsection{Ablation Study}
\subsubsection{Effectiveness of Each Component}
To evaluate the contributions of the proposed technique, we conducted ablation experiments on various combinations of components within the approach, with results summarized in Table \ref{table3}.
Notably, vanilla consistency training struggles to accurately model the distribution of raw motion sequences characterized by dynamic lengths and heterogeneity.
To this end, we compared common variational representations with KL constraints.
We observe that consistency training in variational representation spaces exhibits limitations; however, the proposed bounded finite representation spaces achieve substantial improvements, with the FID metric reduced from 2.837 to 0.615.
Implementing the conditionally guided strategy across both latent representations results in marked performance improvements, thereby validating the versatility of conditionally guided training.
Through the further integration of the clustering guidance module, we enhance consistency training, enabling it to outperform the baseline motion latent diffusion framework in the variational representation space (FID metric of 0.326 vs. 0.473 with MLD) while achieving performance comparable to state-of-the-art methods in the proposed bounded finite state space, thereby underscoring the effectiveness of our approach.

\begin{table*}[!ht] 
    \begin{minipage}{.48\textwidth} 
    \small
        \centering  
        \caption{Ablation study of different guidance scales $\omega$ on the HumanML3D test set. The gray background indicates the default parameters.}
          \setlength{\tabcolsep}{2.8mm}{
           \label{table6}
        \begin{tabular}{cccc}  
            \toprule  
            $\omega$  & \makecell[c]{R-Precision\\ Top-3$\uparrow$} & FID $\downarrow$ & MModality $\uparrow$ \\  
            \midrule  
            1 & $0.806^{\pm .002}$  & $0.250^{\pm .008}$  & $2.958^{\pm .068}$  \\  
            2 & $0.813^{\pm .002}$  & $0.213^{\pm .007}$  & $2.689^{\pm .046}$  \\  
            3 & $0.820^{\pm .003}$  & $0.145^{\pm .004}$  & $2.532^{\pm .064}$  \\  
           \rowcolor{gray!30} 4 & $0.826^{\pm .002}$  & $0.060^{\pm .003}$  & $2.571^{\pm .051}$  \\  
            5 & $0.825^{\pm .002}$  & $0.101^{\pm .009}$  & $2.442^{\pm .066}$  \\  
            \bottomrule  
        \end{tabular}  
        }
    \end{minipage}%
    \hspace{0.4cm}
    \begin{minipage}{.48\textwidth} 
    \small
        \centering  
         \caption{Ablation study of different clustering counts $k$ on the HumanML3D test set. The gray background indicates the default parameters.} 
         \setlength{\tabcolsep}{3mm}{
         \label{table7}
        \begin{tabular}{cccc}  
            \toprule  
            $k$ & \makecell[c]{R-Precision\\ Top-3 $\uparrow$} & FID $\downarrow$& MModality $\uparrow$ \\  
            \midrule  
            256 & $0.823^{\pm .002}$  & $0.129^{\pm .003}$  & $2.537^{\pm .059}$  \\  
            512 & $0.825^{\pm .002}$  & $0.130^{\pm .004}$  & $2.545^{\pm .064}$  \\  
             1024 & $0.823^{\pm .003}$  & $0.113^{\pm .004}$  & $2.567^{\pm .074}$  \\  
           \rowcolor{gray!30} 2048 & $0.826^{\pm .002}$  & $0.060^{\pm .003}$  & $2.571^{\pm .051}$  \\  
            4096 & $0.831^{\pm .002}$  & $0.082^{\pm .002}$  & $2.573^{\pm .067}$  \\  
            \bottomrule 
        \end{tabular} 
        }
    \end{minipage}   
\end{table*}

\begin{figure*}[t]
\centerline{\includegraphics[width=\columnwidth]{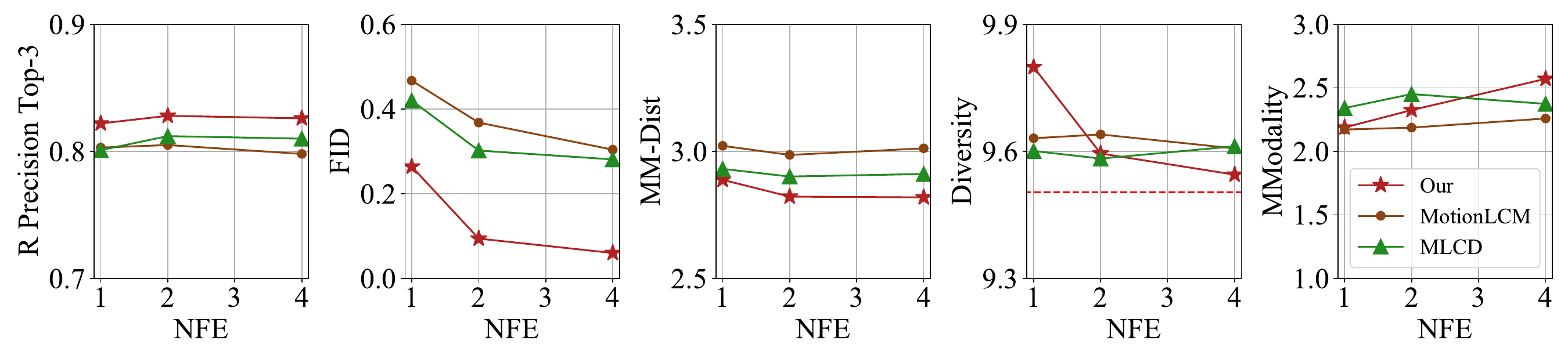}}
\caption{Comparison with latent consistency distillation frameworks, including the latest proposed MotionLCM and ablation experiments of the proposed method in distillation mode.}
\label{fig3}
\end{figure*}
\begin{figure*}[th]
\centerline{\includegraphics[width=\columnwidth]{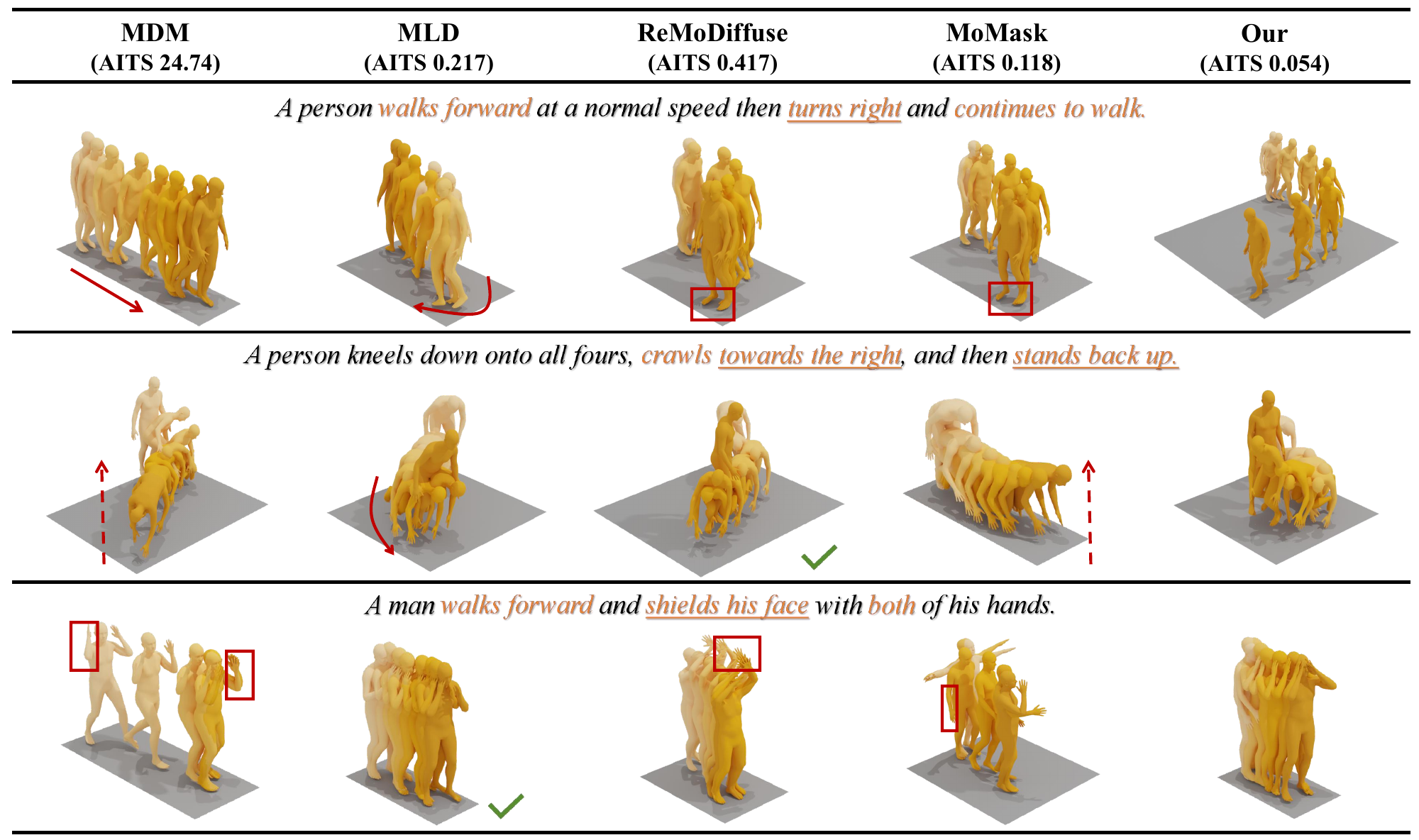}}
\caption{Qualitative analysis of our model and previous models. 
Our model demonstrates improved motion generation performance, matching textual conditions with lower inference costs.
The color of humans darkens over time.}
\label{fig4}
\end{figure*}

\subsubsection{Ablation study on the different model hyperparameters} 
We conduct ablation experiments on motion encoding, varying the token count $n$ and quantization levels $l$, with results presented in Tables \ref{table4} and \ref{table5}.
Unlike MLD, where increasing the token count proves less effective, our framework shows significant improvements in generation quality as the token count increases.
Experimentally, lower quantization levels $l$ lead to a more concise solution space but negatively affect the reconstruction performance of the decoder. As the token count increases, the conciseness achieved by lower quantization levels is balanced against the need for better reconstruction performance.
For the guidance scale $\omega$, we demonstrate the test results in Table \ref{table6}.
Refer to Table \ref{table3} for the results without using conditionally guided consistency training.
We observed that different levels of the guidance scale positively impact generation quality. Notably, when $\omega$ equals 1, the FID metric decreases by 0.292 (FID metric from 0.542 to 0.250). 
As the guidance scale $\omega$ increases, controllability improves gradually, accompanied by a corresponding decrease in diversity. 
This is consistent with previous experience with CFG techniques in diffusion inference.
For the clustering categories number $k$, we show the ablation experiment results in Table \ref{table7}.
The primary contribution of the clustering guidance module is to optimize the FID metrics further. 
Experimental results show that increasing the value of $k$ to 2048 provides finer guidance, leading to a reduction of 0.069 in the FID metric compared to when $k$ is set to 256 and a reduction of 0.15 compared to the baseline without the clustering guidance module (see Table \ref{table3}).
Compared to FID, the effect of $k$ values on controllability and diversity has reached a saturation point. 
While larger $k$ values improve the R-Precision metric, the gain is minimal compared to the improvements achieved through representation designs and conditional trajectory guidance.

\subsection{Comparisons to Consistency Distillation}
One motivation for this paper is to enhance latent consistency training to achieve performance that matches or exceeds traditional latent consistency distillation. 
To this end, we compare our approach with our concurrent work, MotionLCM \cite{Dai2024}, which adheres to the consistency distillation framework. 
The test results are presented in Figure \ref{fig3}. 
Our approach consistently outperforms MotionLCM regarding controllability, generation quality, and diversity under the same NFE. 
It is worth noting that MotionLCM employs pelvic control, i.e., it requires previous awareness of the accurate pelvic trajectory guidance, even during testing and inference.
Considering the differences in detail between the two approaches, we implemented latent consistency distillation with quantized representation and clustering guidance, referred to as MLCD, with results also depicted in Figure \ref{fig3}. 
The experiments demonstrate that our proposed enhancement techniques contribute to the consistency distillation. 
The advantage of consistency training lies in its independence from the performance of the pre-trained model, allowing it to exhibit more significant potential. 
Additionally, it avoids the costs associated with pre-training, reducing both computational and time overhead in training.

\subsection{Qualitative Results}\label{sec5.5}
\begin{figure}[t]
\centerline{\includegraphics[width=0.5\columnwidth]{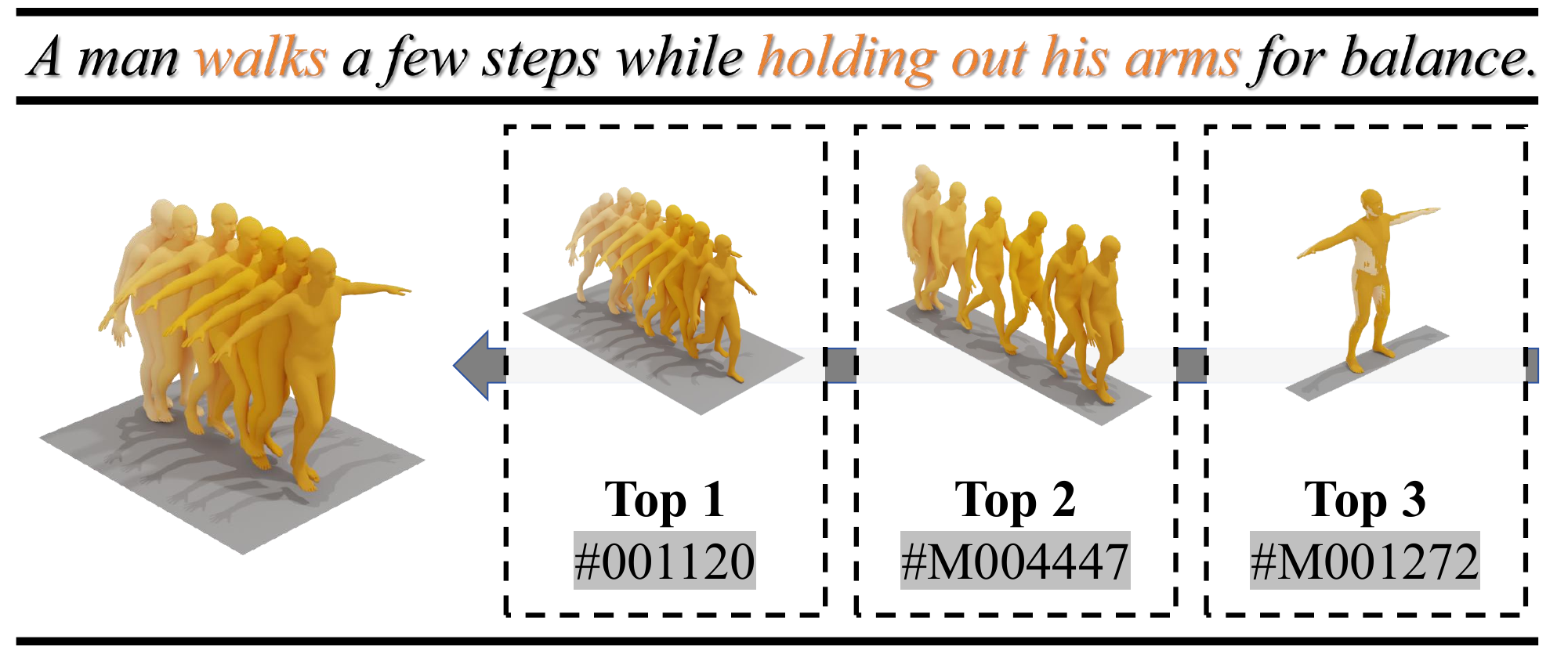}}
\caption{Motion visualizations are sampled from the first three relevant clustering categories, with the \# symbol indicating the unique ID of the sample in the training set.}
\label{fig5}
\end{figure}

We provide a qualitative analysis of our approach, comparing it with two baselines (MDM and MLD) and two state-of-the-art models (ReMoDiffuse and MoMask) in Figure \ref{fig4}. 
While previous works effectively capture the general semantics of instructions, they often fail to account for finer details, such as the need for continuing to walk in the first instructions. 
In contrast, our approach generates fine-grained, high-quality motions with reduced inference time.
Figure \ref{fig5} presents intuitive examples of samples drawn from the first three clustering categories. 
The clustering module provides the denoising process with references to the most relevant samples in the training set.

\begin{figure}[t]
\centerline{\includegraphics[width=0.5\columnwidth]{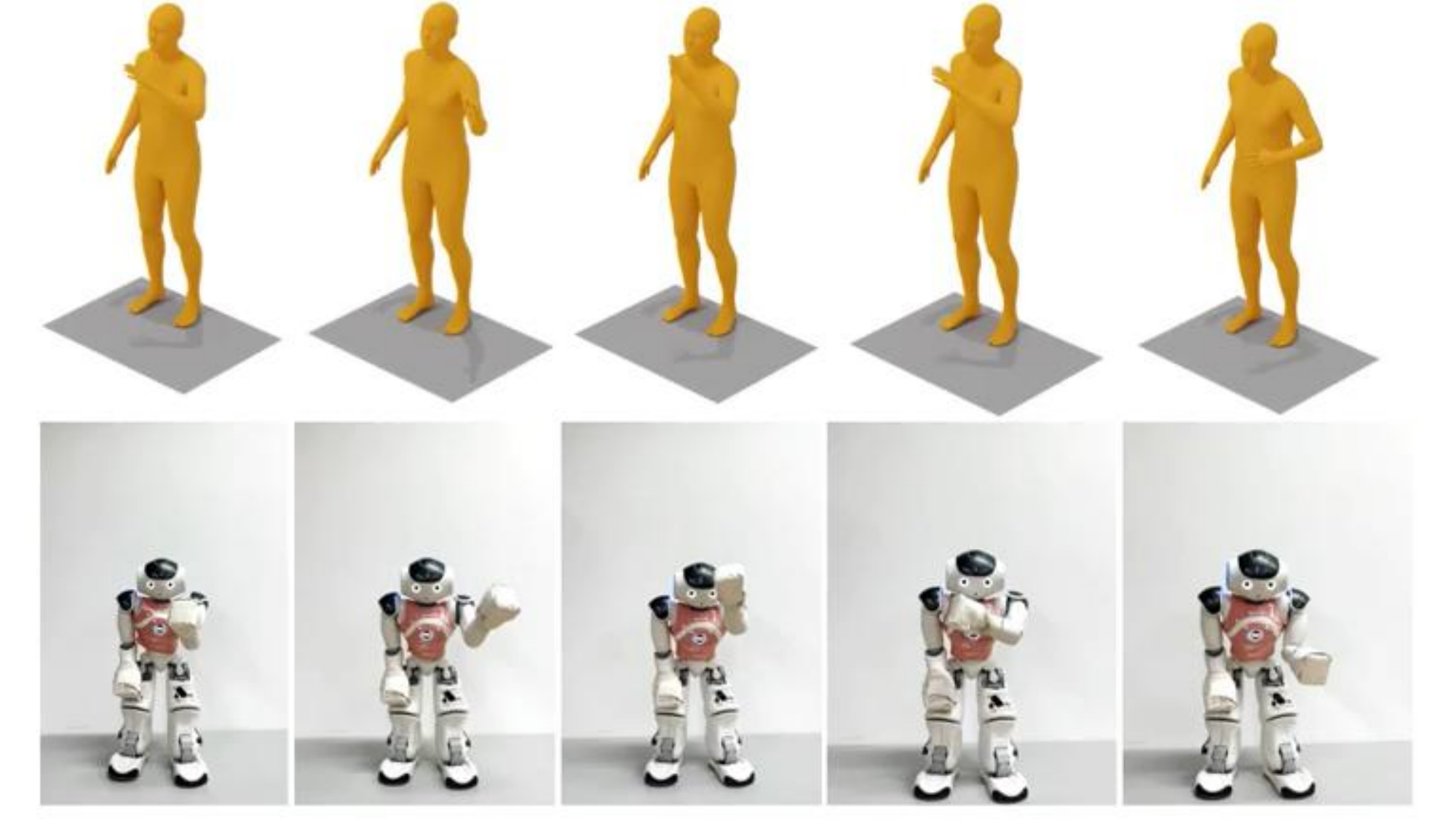}}
\caption{Simple case of the generated motion sequences mapped to the NAO robot through imitation learning. The instruction is that a person waves his left hand.}
\label{fig6}
\end{figure}

\subsection{Inference costs of each component}
To assist readers in evaluating the efficiency of each component, we measured the average inference time for each: 0.0186 seconds for the text encoder, 0.0008 seconds for the clustering guidance module, 0.0071 seconds for the denoiser, and 0.0042 seconds for the motion decoder. 
Notably, text encoding is relatively time-consuming, and the time cost associated with clustering guidance during the inference process is minimal.

\begin{table}[t]
\captionsetup{skip=0pt}
\begin{center}
\caption{User studies for quantitative comparison. }
    \begin{tabular}{p{3.5cm}C{1.8cm}C{1.8cm}}
    \toprule
    Methods Compared & \makecell[c]{Fidelity} & \makecell[c]{Condition \\ Matching}\\
    \midrule
    MLCT vs. MLD & 57.6\% & 60.4\%\\
    MLCT vs. ReMoDiffuse & 55.4\% & 58.1\%\\
    MLCT vs. Ground Truth & 47.9\% & 48.2\% \\
    \bottomrule
    \end{tabular}
  \label{tab4}%
\end{center}
\end{table}

\subsection{User Study}\label{SectionE}
Similar to previous work \cite{Chen2023}, this paper conducts a user study. 
We randomly sampled 30 sets of text descriptions from the HumanML3D test set, with MLCT, MLD, and ReMoDiffuse generating corresponding motions.
Forty-six participants were invited to compare MLCT with MLD, ReMoDiffuse, and the ground truth motion. 
We employed a forced-choice paradigm, asking participants, "Which of the two motions is more realistic?" and "Which of the two motions corresponds better to the text prompt?" 
As shown in Table \ref{tab4}, our method outperforms MLD and ReMoDiffuse with a minimal inference cost of 4 NFEs and demonstrates competitive fidelity with ground truth motions.

\section{Discussion}
The proposed framework enables fast, high-quality motion sampling, offering cross-modal solutions for human-centered interaction applications. 
As an example, we briefly explore its application to humanoid robot imitation learning, as illustrated in Figure \ref{fig6}. 
The generated motion features, converted into joint sequences, can guide the motion planning of humanoid robots based on reinforcement learning pre-training. 
Additionally, the generated motion can be applied in games or movies, allowing virtual characters to respond to user instructions in real-time.

Although our method demonstrates progress, there are still areas for improvement:
(i) The MLCT follows the diffusion framework, which encourages diversity due to its stochastic nature but may occasionally produce undesired results. Moreover, our framework directly learns distributions from data without incorporating physical laws, a limitation also present in previous work, such as ReMoDiffuse.
(ii) Our set of textual instructions focuses on the annotated data of HumanML3D, but it may be limited, and out-of-domain instructions may result in unreasonable sample generation.

We would like to incorporate more physical constraints to minimize undesired motion generation.
Furthermore, with the emergence of large language models, future research could explore their potential to understand a broader context of semantic instructions.

\section{Conclusion}
This paper presents a motion latent consistency training framework designed for fast, high-fidelity, text-matched motion generation. This framework encodes human motion sequences into tokens through the quantization constraint, ensuring bounded finite states that optimize the latent representation. Additionally, we introduce a conditionally guided consistency training framework and a clustering guidance module to enhance conditional controllability and provide supplementary references for solution distribution. Our model and its components have been validated through extensive experiments, demonstrating an optimal trade-off between performance and computational efficiency with minimal NFE. This approach provides a valuable reference for training subsequent latent consistency models across various tasks.

\bibliographystyle{IEEEtran}
\bibliography{references.bib}

\end{document}